\documentclass[runningheads]{llncs}
\usepackage{graphicx}
\usepackage{amssymb}
\usepackage{subcaption}
\usepackage{wrapfig}
\usepackage{booktabs}
\usepackage[misc]{ifsym}
\usepackage{hyperref,xcolor}

\usepackage{subfiles}
\begin{document}
%
\title{Blackbox Adaptation for Medical Image Segmentation }
%
%
%
%

%
\author{Jay N. Paranjape\inst{1,}\Letter \and
Shameema Sikder\inst{2,3} \and
Vishal M. Patel\inst{1} \and
S. Swaroop Vedula\inst{3}
}
\authorrunning{J. Paranjape et al.}
\institute{Department of Electrical and Computer Engineering, The Johns Hopkins University, Baltimore, USA \\
\email{jparanj1@jhu.edu}
\and
Wilmer Eye Institute, The Johns Hopkins University, Baltimore, USA \and Malone Center for Engineering in Healthcare, The Johns Hopkins University, Baltimore, USA
}
\maketitle              
%
\begin{abstract}
In recent years, various large foundation models have been proposed for image segmentation. These models are often trained on large amounts of data corresponding to general computer vision tasks.  Hence, these models do not perform well on medical data. There have been some attempts in the literature to perform parameter-efficient finetuning of such foundation models for medical image segmentation. However, these approaches assume that all the parameters of the model are available for adaptation. But, in many cases, these models are released as APIs or blackboxes, with no or limited access to the model parameters and data. In addition, finetuning methods also require a significant amount of compute, which may not be available for the downstream task. At the same time, medical data can't be shared with third-party agents for finetuning due to privacy reasons. To tackle these challenges, we pioneer a blackbox adaptation technique for prompted medical image segmentation, called BAPS. BAPS has two components - (i) An Image-Prompt decoder (IP decoder) module that generates visual prompts given an image and a prompt, and (ii) A Zero Order Optimization (ZOO) Method, called SPSA-GC that is used to update the IP decoder without the need for backpropagating through the foundation model. Thus, our method does not require any knowledge about the foundation model's weights or gradients. We test BAPS on four different modalities and show that our method can improve the original model's performance by around 4\%. The code is available at \url{https://github.com/JayParanjape/Blackbox}.


\end{abstract}

\section{Introduction}
\begin{figure}
\centering
\includegraphics[width=\textwidth]{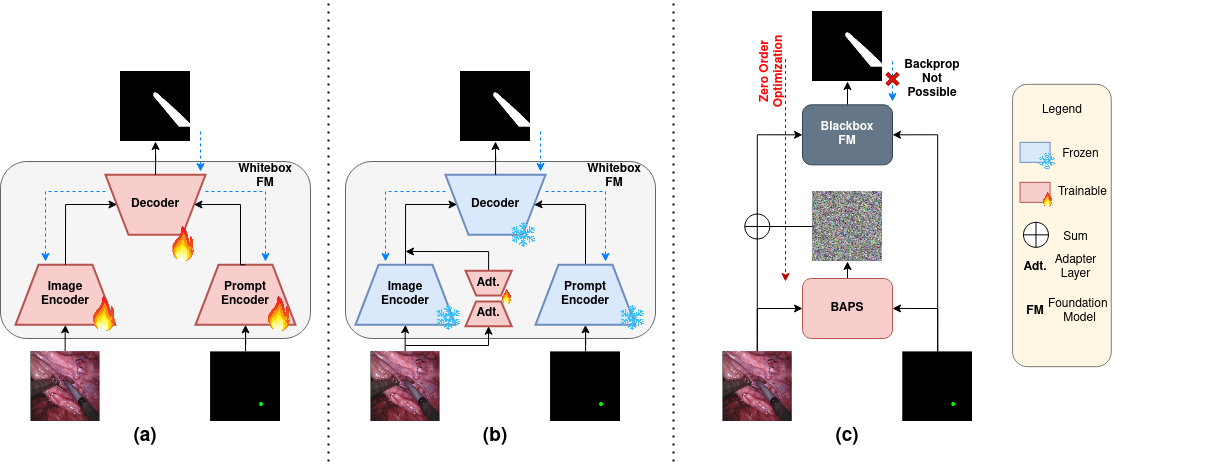}
\caption{Comparison of our method (c) against full finetuning of a Foundation Model (FM), (a) and common adaptation techniques (b).} \label{intro_pic}
\end{figure}

Image segmentation is a fundamental problem in medical image analysis tasks. Many deep learning-based approaches have been proposed in the literature that segment out various regions of interest across different medical modalities \cite{dl5_survey}. In recent years, many foundation models have been proposed, which are large models with billions of parameters that show excellent performance on downstream tasks like classification \cite{clip} and segmentation \cite{sam,medsam,sam_msft}. Following the success of these foundation models, various adaptation approaches have been proposed that transfer the knowledge learnt by these large-scale models to medical segmentation \cite{sam_survey}. These adaptation methods are commonly termed as Parameter Efficient Finetuning (PEFT) methods and they aim to utilize a given foundation model while tuning only a fraction of its parameters. However, all PEFT methods make two over-optimistic assumptions. First, they assume that all parameters of the foundation model are available during training. However, it often occurs that companies release their AI models as APIs or blackboxes instead of releasing the entire parameter set, training dataset, or codebase due to proprietary concerns. In such conditions, all the PEFT methods would fail since they require gradient computation to work. Secondly, many PEFT methods assume the availability of high compute resources \cite{medsamadapter,samed}, which is not realistic. At the same time, the transfer of medical data to a third-party system with more resources would result in privacy concerns. In this work, we tackle these concerns by pioneering a blackbox adaptation method for medical image segmentation called Blackbox Adapter for Prompted Segmentation (BAPS). BAPS uses a frozen image encoder and a trainable Image-Prompt Decoder (IP Decoder), to produce an input-dependent per-pixel prompt. This is added to the original image and provided to the foundation model blackbox. We train the IP Decoder using a recently proposed ZOO method called Simultaneous Perturbation Stochastic Approximation with Gradient Correction (SPSA-GC) \cite{blackvip} which does not require computation of any gradient, thus, not needing any parameter information about the foundation model. The IP Decoder is a lightweight module and hence, our method requires minimal compute resources. In summary, our contributions can be listed as follows: \\
\noindent\textbf{1)} To the best of our knowledge, this is the first paper to explore blackbox adaptation for medical image segmentation. For this, we propose BAPS which has a frozen pre-trained encoder and a lightweight decoder module that can be trained using derivative-free optimization methods. A visual comparison of our method with existing approaches can be seen in Figure \ref{intro_pic}.\\
\noindent\textbf{2)} We show that BAPS can improve the original foundation model's performance on four widely used public datasets of different modalities, including endoscopic images, dermoscopic images, gastrointestinal polyp images, and retinal images. For this purpose, we conduct experiments using two recently proposed popular image segmentation foundation models - SAM \cite{sam} and MedSAM \cite{medsam}.

\section{Related Work}
The motivation for using blackbox adaptation arose from the Natural Language Processing (NLP) field. Many custom AI models available in the NLP industry are deployed as a proprietary service or API \cite{gpt}. Hence, several works attempt to adapt them for customized datasets using ZOO algorithms as the model weights and gradients are not available \cite{bbt,bbtv2,rlprompt}. BBT \cite{bbt} and BBTv2 \cite{bbtv2} use a ZOO method called Covariance Matrix Adaptation Evolution Strategy (CMA-ES) \cite{cmaes1,cmaes2}, while RLPrompt \cite{rlprompt} uses a reinforcement learning gradient-free based method to tune prompts, which are then added to the input to the foundation model. However, reinforcement learning and evolutionary strategy-based methods tend to show high variance and are unstable \cite{cmaes_fail}. Further, they do not work as well for vision tasks \cite{blackvip}. There are very few methods that perform blackbox adaptation for vision tasks \cite{bar,blackvip}. BAR \cite{bar} uses a one-sided approximation for gradient estimation. However, this was found to be inaccurate empirically by a succeeding approach called BlackVIP \cite{blackvip}. BlackVIP uses a two-sided approximation for gradient estimation and proposes a different ZOO method for tuning weights. However, both BAR and BlackVIP are proposed only for the classification task with CLIP \cite{clip} as the blackbox foundation model. In this work, we pioneer blackbox adaptation for the recently proposed task of prompted segmentation.
\section{Proposed Method}
\begin{figure}
\centering
\includegraphics[width=.85\textwidth]{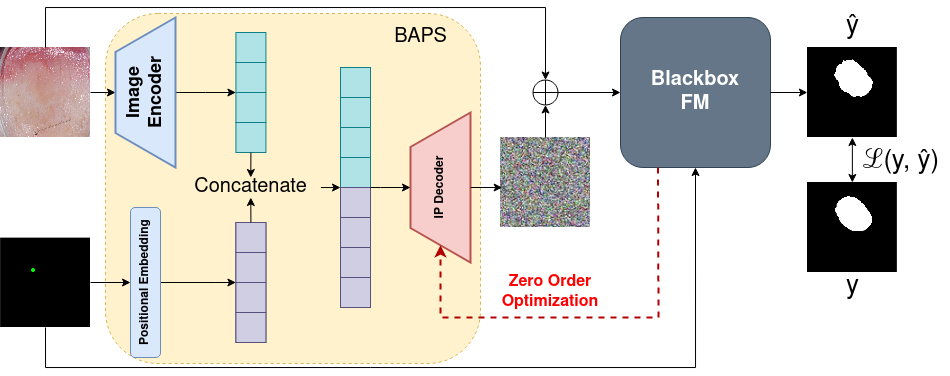}
\caption{Overview of BAPS.} \label{arch}
\end{figure}

\noindent \textbf{Model Architecture: }
Foundation models for segmentation usually perform the task of prompted segmentation. Given an input image and a prompt, the foundation model produces a segmentation mask that corresponds to the given prompt. Hence, for BAPS, we consider an image and a point prompt as the inputs. The overview of BAPS is shown in Figure \ref{arch}. The input image is passed through a pre-trained image encoder that produces image embeddings. We use the Vision Transformer (ViT) encoder with Masked Autoencoder (MAE) pretraining \cite{vit_mae} as our image encoder because of its strong innate understanding of images. Hence, it generates highly representative features. On the other hand, the point prompt is converted into positional embeddings based on its relative position in the image. This uses a sinusoid function for generating the embeddings, similar to ViT \cite{vit}.

The image and prompt embeddings are then concatenated and passed to a module called the Image-Prompt Decoder (IP-Decoder). This module is a deconvolution network that generates a visual prompt as its output. This is then added to the original image and passed to the blackbox foundation model along with the original point prompt. Thus, the job of the IP-Decoder is to learn a residual visual prompt that when added to the original image, will make it easier for the foundation model to segment out the correct shape. Please note that the IP-Decoder is the only trainable module in BAPS.

Once the blackbox model generates the prediction, it is compared with the label using a sum of the Binary Cross Entropy (BCE) loss and dice loss \cite{dice_loss}. 
However, note that the gradients cannot be backpropagated since the weights in the blackbox model are not available. Hence, we use a ZOO method called SPSA-GC \cite{blackvip}, which we describe next.
\\

\noindent \textbf{SPSA-GC: }
Simultaneous Perturbation Stochastic Approximation (SPSA) \cite{spsa} is a Zero Order Optimization (ZOO) technique that estimates gradients effectively using a two-sided approximation of the derivative. Given a set of parameters \(\phi\) and a loss function \(\mathbb{L}\), the estimated gradients \(\hat{g}\) are given as follows:
\begin{equation}
    \hat{g_i} = \frac{\mathbb{L}(\phi_{i} + c\Delta_i) - \mathbb{L}(\phi_{i} - c\Delta_i)}{2c}\Delta_i^{-1},
\end{equation}
where \(c \in [0,1]\) is a hyperparameter and \(\Delta\) represents a random perturbation vector of the same shape as \(\phi\). Each element of \(\Delta\) is sampled uniformly from \([-1,-0.5] \cup [0.5,1]\). Here, \(i\) represents the iteration number. This estimate of the gradient can be used to update the weights of the model as \(\phi_{i+1} \leftarrow \phi_i - \alpha\hat{g_i}\), where \(\alpha\) is the learning rate. While SPSA is a good approximator, it has been shown that it can lead to slower convergence \cite{blackvip}. Hence, in SPSA-GC a momentum term $m$ is added while updating the weights as follows:
\begin{equation}
    m_{i+1} = \beta m_i - \alpha \hat{g_i}(\phi_i + \beta m_i),
    \;\;\;\;
    \phi_{i+1} = \phi_i + m_{i+1},
\end{equation}
where $\beta$ denotes the weight of the momentum. This allows the IP Decoder to train quicker and more stably. However, we found that given a point prompt, the foundation model produces a reasonable mask without any sort of optimization (zero-shot performance). Thus, the approximated gradients are small in magnitude and the system gets stuck at local minimas more often. To alleviate this, as an implementation detail, we add a mechanism to detect the local minima and increase the learning rate for some iterations during training to help the model get out of the local minima. This is detailed in the supplementary document.

\section{Experiments and Results}
We choose the recently proposed Segment Anything Model (SAM) \cite{sam} and MedSAM \cite{medsam} as the blackbox foundation models for our experiments. SAM and MedSAM are foundation models for the task of promptable segmentation. Given an image and a prompt in the form of point, text, mask or bounding box, they can segment out the object of interest corresponding to the prompt. SAM is trained on a large corpus of natural images while MedSAM is tuned on medical data including 3D and 2D images. Majority of the training data of MedSAM includes CT and MRI images. We use the point-based prompt for our experiments. Thus, for a given image and a point prompt, we pass it through BAPS and generate a visual prompt. This is added to the original image and sent to the blackbox foundation model along with the point prompt. We evaluate our method on four different datasets and calculate the Dice Score (DSC) and Hausdorff Distance at the 95th percentile (HD95). For our baselines, we consider the zero-shot performance of the foundation model without any adaptation, and Visual Prompting (VP) \cite{vpt} with SPSA-GC. We measure the upper bound of the segmentation performance using a white box adaptation of SAM and MedSAM using LoRA \cite{lora}, similar to various SAM-based adaptation methods \cite{medsamadapter,samed}.\\

\noindent {\bf{Datasets.}} We use four widely used publicly available datasets for our experiments. Kvasir-Seg \cite{kvasirseg} consists of gastrointestinal polyp images, divided into 600 training, 100 validation, and 300 test images.  ISIC2018 \cite{isic} contains dermoscopic images for skin lesion segmentation. It is divided into 2594 images for training, 100 for validation, and 100 for testing. The third dataset is REFUGE \cite{refuge}, which has retinal images for optic disk and optic cup segmentation. It is divided into 800 training, 800 validation, and 800 testing images. The fourth dataset is Endovis 17 \cite{ev17}, which has images of endoscopic surgery. It is further divided into 2878 testing and 3231 training images, out of which we use 366 for validation. The dataloaders will be made available along with the code after the review process.\\

\noindent {\bf{Experimental Setup.}}
For all the datasets, we use the same set of data augmentations. These include random rotation up to 10 degrees, random brightness changes with scale 2, and random saturation changes with scale 2. The images are scaled to the resolution $512 \times 512$. Based on the validation set performance of these datasets, we set hyperparameters of the model as: $c = 0.01$, $\alpha = 0.005$ for Endovis17 and $\alpha = 0.01$ for other datasets,  $\beta = 0.9$. For all datasets, we use a batch size of 32. Training is done on a single Nvidia RTX A5000 GPU and uses only 6GB of memory. Note that in our case, we also run the blackbox on the same GPU. However, in practice, training would be cheaper since the forward propagation in the blackbox will occur at the server. The blackboxes SAM and MedSAM both are initialized using their ViT-base checkpoints.

\begin{figure}
\centering
\includegraphics[width=0.7\textwidth]{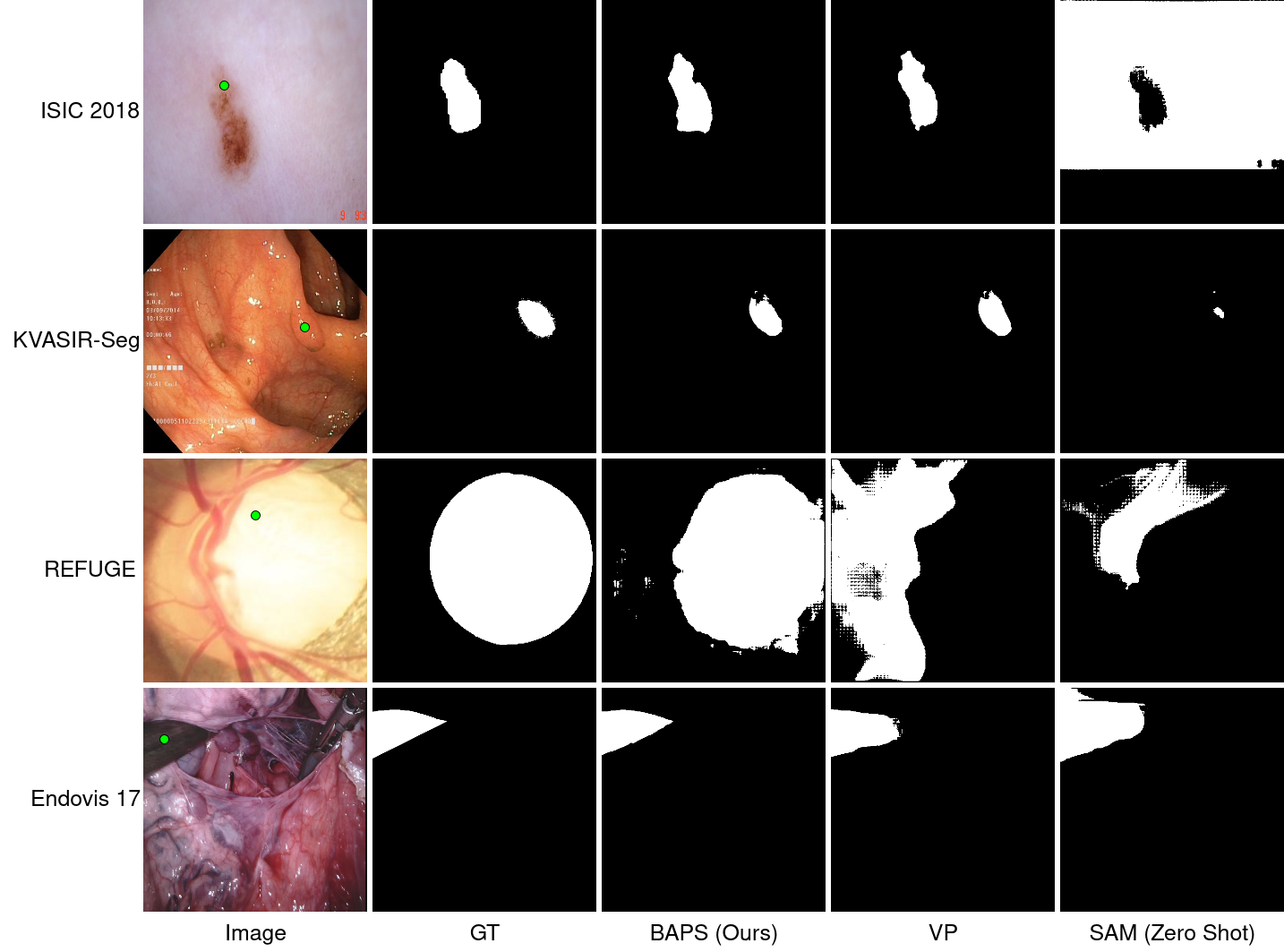}
\caption{Qualitative Results on all the datasets. GT - ground truth, VP - visual prompting \cite{vpt}. The green dot in the image denotes the point prompt given to the blackbox foundation model} \label{results}
\end{figure}

\begin{table}
\centering
\caption{Our approach improves over the zero-shot performance of the blackbox model SAM (row 1). Having the IP Decoder improves performance over direct visual prompt tuning (row 2). The performance of our method is upper bounded by the white box adaptation methods like LoRA (row 4).}
\label{tab_results}
\resizebox{\columnwidth}{!}{
\begin{tabular}{|c|c|c|c|c|c|c|c|c|}
\hline
\multicolumn{9}{|c|}{\textit{Foundation Model - SAM}}\\
\hline
&\multicolumn{2}{|c|}{Kvasir-Seg} & \multicolumn{2}{|c|}{ISIC2018} & \multicolumn{2}{|c|}{REFUGE} & \multicolumn{2}{|c|}{Endovis 17}\\
\hline
Method &  DSC ($\uparrow$) & HD95 ($\downarrow$) & DSC ($\uparrow$) & HD95 ($\downarrow$) & DSC ($\uparrow$) & HD95 ($\downarrow$) & DSC ($\uparrow$) & HD95 ($\downarrow$)\\
\hline
SAM (ZS) \cite{sam} & 0.68 & 99.3 & 0.66 & 85.98 & 0.38 & 214.98 & 0.60 & 89.46\\
VP \cite{vpt} & 0.69 & 99.3 & 0.70 & 70.96 & 0.39 & 201.67 & 0.63 & 83.77 \\
BAPS (Our Approach) & \textbf{0.72} & \textbf{83.55} & \textbf{0.74} & \textbf{70.3} & \textbf{0.44} & \textbf{176.67} & \textbf{0.65} & \textbf{81.56} \\
\hline
Whitebox LoRA \cite{lora} & \textit{0.88} & \textit{22.23} & \textit{0.85} & \textit{30.95} & \textit{0.85} & \textit{19.22} & \textit{0.68} & \textit{45.39} \\
\hline
\end{tabular}
}
\end{table}
\noindent {\bf{Results.}}
We tabulate the quantitative results on each of the datasets in Table \ref{tab_results} for SAM and in Table \ref{tab_results2} for MedSAM. For all four datasets, we see an average of 5\% improvement in the Dice Score with our method over the foundation model SAM and an average of 7\% over MedSAM. For all the results, evaluation is done five times with randomly selected point prompts and the mean value is listed in the table. The standard deviation in each case is less than 0.01. The increase in performance can be attributed to the strong pre-trained encoder and IP Decoder of BAPS, which generates a visual prompt as a function of the input image and the point prompt. Some samples of the modified images after adding the visual prompt are shown in Figure \ref{vp}. In row 2, we compare our method with simply adding the same visual prompt for each image-point pair (no encoder or I-P Decoder). We see significant improvement, showing the effectiveness of the encoder-decoder structure. This can also be seen in supplementary Figure 1, where we plot the training progress of BAPS in comparison to VP, with MedSAM as the blackbox for ISIC2018 and REFUGE. For both these cases, we see that the average error for BAPS decreases consistently with the number of iterations. All results with BAPS have a p-value of at most $10^{-8}$. Qualitative results are shown in Figure \ref{results}.\\

\begin{table}
\centering
\caption{Our approach improves over the zero-shot performance of the blackbox model MedSAM (row 1). Having the IP Decoder improves performance over direct visual prompt tuning (row 2). The performance of our method is upper bounded by the white box adaptation methods like LoRA (row 4).}
\label{tab_results2}
\resizebox{\columnwidth}{!}{
\begin{tabular}{|c|c|c|c|c|c|c|c|c|}
\hline
\multicolumn{9}{|c|}{\textit{Foundation Model - MedSAM}}\\
\hline
&\multicolumn{2}{|c|}{Kvasir-Seg} & \multicolumn{2}{|c|}{ISIC2018} & \multicolumn{2}{|c|}{REFUGE} & \multicolumn{2}{|c|}{Endovis 17}\\
\hline
Method &  DSC ($\uparrow$) & HD95 ($\downarrow$) & DSC $(\uparrow$) & HD95 ($\downarrow$) & DSC ($\uparrow$) & HD95 ($\downarrow$) & DSC ($\uparrow$) & HD95 ($\downarrow$)\\
\hline
MedSAM (ZS) \cite{medsam} & 0.68 & 98.4 & 0.70 & 87.23 & 0.36 & 214.29 & 0.63 & 87.52\\
VP \cite{vpt} & 0.68 & 90.36 & 0.71 & 99.2 & 0.36 & 214.48 & 0.63 & 85.58 \\
BAPS (Our Approach) & \textbf{0.72} & \textbf{80.01} & \textbf{0.79} & \textbf{66.6} & \textbf{0.44} & \textbf{168.07} & \textbf{0.65} & \textbf{82.4} \\\hline
Whitebox LoRA \cite{lora} & \textit{0.90} & \textit{20.05} & \textit{0.86} & \textit{30.1} & \textit{0.85} & \textit{19.15} & \textit{0.68} & \textit{42.92} \\
\hline
\end{tabular}
}
\end{table}

\section{Ablation Analysis}

\noindent {\bf{Ablation Over Pretrained Image Encoder.}} 
We use the MAE pre-trained ViT image encoder from Meta (ViT-MAE) \cite{vit_mae} since it gives a stronger featurization of images over other encoders. To verify this claim, we make an ablation analysis by changing the pre-trained encoder and measuring the performance on the ISIC 2018 dataset. The results are shown in Table \ref{ablation_enc}, where we compare the results of ViT-MAE with three popular image encoders. We find that ViT-MAE outperforms CLIP \cite{clip} and DINO-Resnet50 \cite{dino_resnet50} significantly. ViT \cite{vit} is a strong encoder. However, the MAE pretraining further improves the downstream performance by 2\%
\\
\noindent {\bf{Ablation Over SPSA.}} 
We empirically test the effectiveness of SPSA-GC as a ZOO method by removing each of the components of the algorithm and measuring performance on the ISIC 2018 dataset. As seen in Table \ref{ablation_spsa}, we start with the zero-shot performance of SAM \cite{sam}, which has no required optimization. Using just SPSA \cite{spsa} with zero momentum increases this performance by only 1\%. This is further improved by adding a momentum of 0.9 as suggested by SPSA-GC \cite{blackvip}, finally giving a DSC of 0.74.

\begin{table}
\parbox{.45\linewidth}{
\centering
\caption{Ablation analysis of different pre-trained image encoders on ISIC 2018 dataset}
\label{ablation_enc}
\begin{tabular}{|c|c|}
\hline
Image encoder & DSC \\
\hline
CLIP \cite{clip} & 0.65 \\
DINO-Resnet50 \cite{dino_resnet50} & 0.69 \\
ViT \cite{vit} & 0.72 \\
ViT-MAE \cite{vit_mae} (used in BAPS) & \textbf{0.74} \\
\hline
\end{tabular}
}
\hfill
\parbox{.45\linewidth}{
\centering
\caption{Ablation analysis of component-wise importance of the zero-order optimization method.}
\label{ablation_spsa}
\begin{tabular}{|c|c|}
\hline
Optimization method & DSC\\
\hline
SAM (ZS) \cite{sam} (No ZOO) & 0.66\\
SPSA \cite{spsa} & 0.68\\
SPSA-GC (Ours) & \textbf{0.74}\\
\hline
\end{tabular}
}
\end{table}

\begin{figure}
\centering
\includegraphics[width=.7\textwidth]{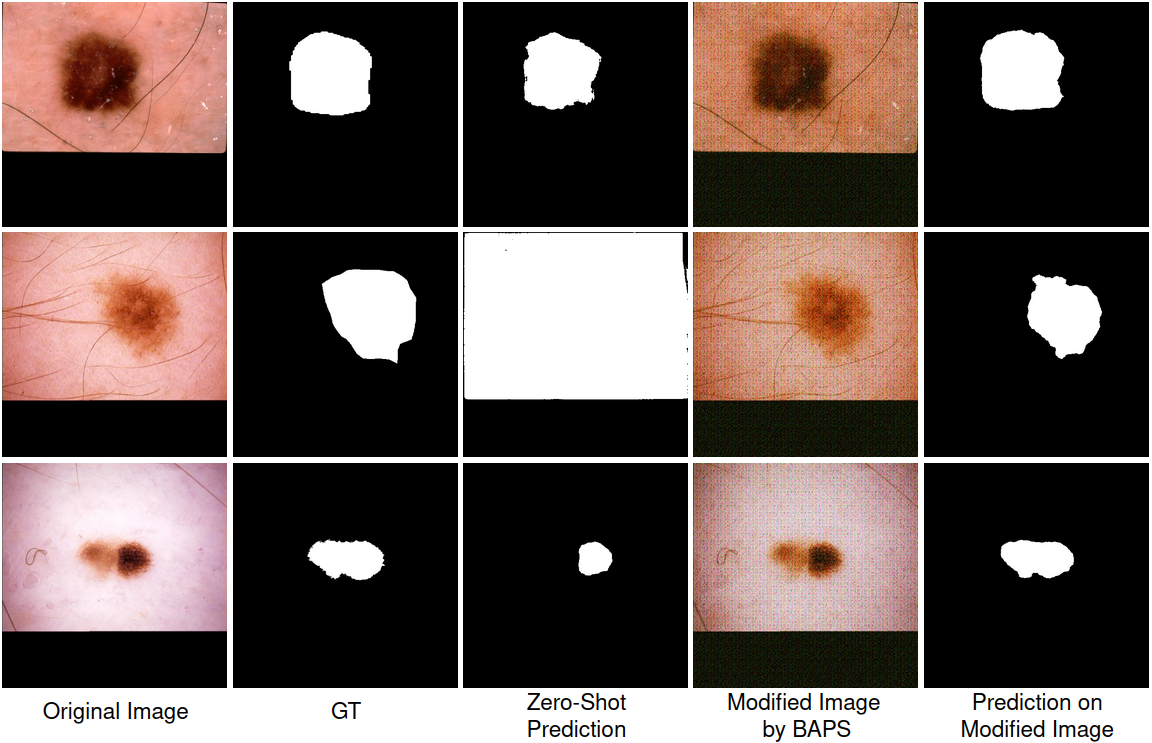}
\caption{Visualizing the effect of adding the visual prompt on ISIC 2018 dataset. The blackbox used for these results is MedSAM.} \label{vp}
\end{figure}

\noindent\textbf{Visualizing the Modified Images: }To test the effectiveness of BAPS, we compare the visual results of the blackbox foundation model MedSAM with and without adding the visual prompt generated by BAPS. These results are shown in Figure \ref{vp}. Here, the zero-shot prediction of MedSAM generates inaccurate masks as seen in column 3. However, after adding the learnt visual prompt to the original image, MedSAM can correctly generate the masks. The modified image can be seen in column 4 of the figure.

\noindent {\bf{Memory Analysis.}} We perform an analysis of the number of parameters required to be trained by the client and the memory requirement of the generated checkpoint. Here, the client refers to the machine used at the downstream application level. Note that for our method, SAM is a blackbox and does not have to be run on the client machine, resulting in significantly lower memory consumption. This is one of the advantages of the proposed blackbox paradigm. BAPS requires just one hundred thousand parameters to be tuned, thus making the memory required to store the model checkpoint only 0.4 MB as compared to hundreds to thousands of MBs required by whitebox adaptation methods, which have at least ten times as many parameters.

\section{Conclusion}
In this work, we proposed one of the first blackbox adaptation methods, called BAPS, for the adaptation of foundation models for prompted segmentation. BAPS consists of a pretrained image encoder and a trainable IP decoder, that generates a visual prompt as a function of the input image and given prompt. This visual prompt is added to the original image and given to the foundation model. The IP decoder is trained using a novel Zero Order Optimization (ZOO) method called SPSA-GC. We test BAPS on four different public datasets in medical segmentation and verify its effectiveness for the recently proposed and popular foundation models SAM and MedSAM. Finally, we carry out an ablation study to gauge the effectiveness of different design decisions of BAPS. Thus, our proposed method can efficiently perform blackbox adaptation of SAM without the requirement of gradients.

%
%
%
\bibliographystyle{splncs04}
\bibliography{bibliography_v2}

\end{document}


\title{Supplementary - Blackbox Adaptation for Medical Image Segmentation}

%
%
%
%

%
\author{Anonymous MICCAI submission}
\institute{Paper ID 668}
\maketitle
\begin{figure}
\centering
\includegraphics[width=.85\textwidth]{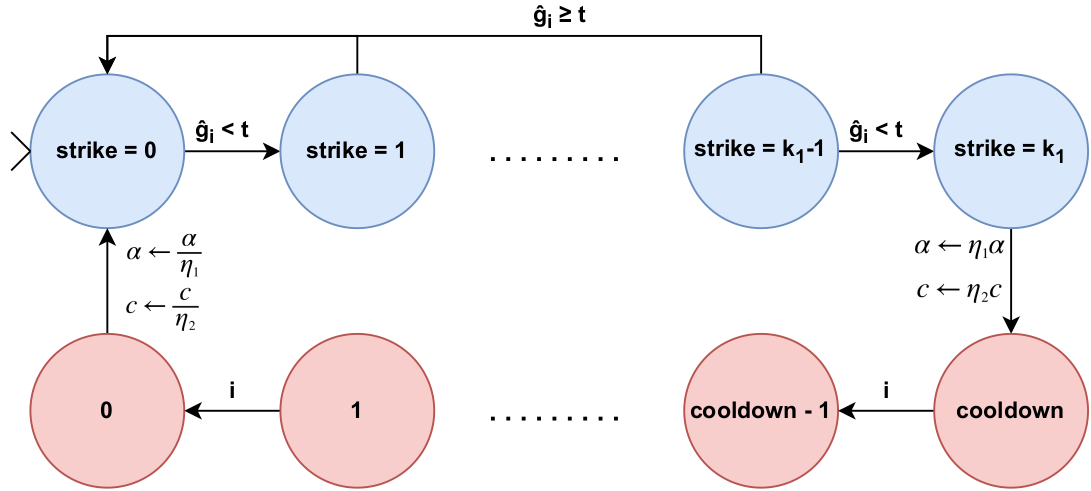}
\caption{Overview of SPSA-GEASS. The system starts at \(strike=0\). If the estimated gradient is greater than the threshold, \(strike\) increases, or else the system reverts to the original state. If \(strike\) reaches \(k_1\), the learning rate and perturbation step parameter increase significantly. Then, the \(cooldown\) reduces every iteration until 0. After this, the system returns to its initial state.} \label{geass}
\end{figure}

            
\noindent\textbf{SPSA-GEASS: }Given a point prompt, the foundation model produces a reasonable mask without any sort of optimization (zero-shot performance). Thus, the approximated gradients are small in magnitude and the system tends to get stuck at local minimas. In order to alleviate this, we propose a small modification to the SPSA-GC algorithm called SPSA-Gradient Enhancement Assistance (SPSA-GEASS). As shown in Figure \ref{geass}, we define two parameters \(strike\) and \(cooldown\). \(strike\) is initialized to zero during the start of the training. During training, if the magnitude of gradient \(\hat{g_i}\) falls below a threshold, \(strike\) is incremented until it reaches a limit \(k_1\). After this, the learning rate \(\alpha\) and the perturbation step parameter \(c\) in SPSA-GC are multiplied by hyperparameters \(\eta_1\) and \(\eta_2\) respectively, for a number of iterations equal to \(cooldown\). This emulates the effect of taking a large step so that the model might move out of a local minima. The parameter \(strike\) determines the tolerance of low-magnitude gradients during training, while the parameter \(cooldown\) controls the number of iterations for which large steps are taken. However, these parameters ( $strike, cooldown, \eta_1, \eta_2, k_1$ ) require careful selection and would vary across datasets and other hyperparameters like the learning rate.

\begin{figure}
  \centering
  \subfloat[ISIC2018]{\includegraphics[width=0.49\textwidth]{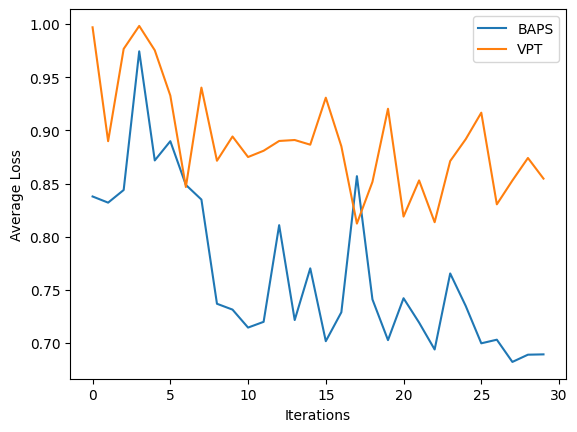}}
  \hfill
  \subfloat[REFUGE]{\includegraphics[width=0.49\textwidth]{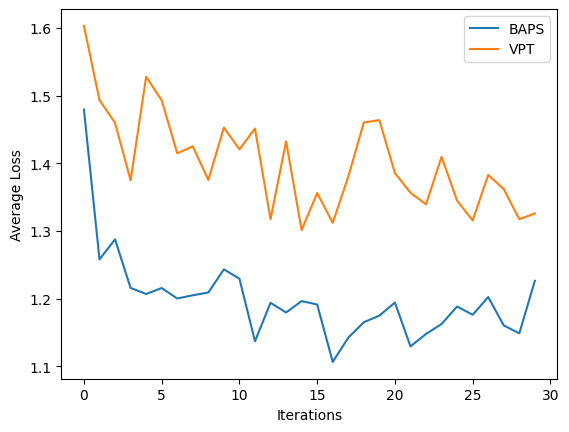}}
  \caption{The training progress of BAPS versus VPT with MedSAM as the blackbox foundation model.}
\label{tr_progress}
\end{figure}

\noindent\textbf{Training Progress of BAPS: }To test the effectiveness of BAPS, we compare its training progress with that of Visual Prompting for two datasets with MedSAM as the foundation model. These plots are shown in Figure \ref{tr_progress}. Here, we see a clear gap between the loss values, with BAPS consistently showing a lower loss value.


